\documentclass[
]{ceurart}

\usepackage{listings}
\usepackage{tabularx}
\usepackage{multirow}
\usepackage{hyperref}
\usepackage{subcaption}
\usepackage{xcolor}

\lstdefinestyle{mystyle}{
    basicstyle=\ttfamily\small,
    breaklines=true,
    showspaces=false,                
    showstringspaces=false,          
    columns=flexible,
    keepspaces=true                  
}

\begin{document}

\copyrightyear{2022}
\copyrightclause{Copyright for this paper by its authors.
  Use permitted under Creative Commons License Attribution 4.0
  International (CC BY 4.0).}

\conference{2nd Causal Neuro-symbolic Artificial Intelligence (Causal NeSy): Toward Agentic LLMs with Neuro-Symbolic and Graph Based Reasoning}

\title{Enhancing Small Language Models Reasoning through Knowledge Graph Grounding}


\author[1]{Dimitrios Kelesis}[%
orcid=0000-0002-3434-2717,
email=dkelesis@iit.demokritos.gr,
]
\fnmark[1]
\cormark[1]

\address[1]{Institute of Informatics and Telecommunications, National Center for Scientific Research ``Demokritos", Athens, Greece}

\author[1]{Konstantinos Bougiatiotis}[%
orcid=0000-0002-1910-2758,
]
\fnmark[1]

\author[1]{Georgios Paliouras}[%
]
\cortext[1]{Corresponding author.}
\fntext[1]{These authors contributed equally.}

\begin{abstract}
 Although large language models (LLMs) have set benchmarks for zero-shot reasoning, their deployment remains cost-prohibitive and environmentally taxing. Small Language Models (SLMs) offer a sustainable alternative, but prone to errors, on tasks requiring complex, multi-hop logical grounding. We investigate a neuro-symbolic agentic framework to enhance the reasoning capabilities of SLMs, specifically Gemma 3 (1B, 4B) and Llama 3.2 (3B), using the CLUTRR kinship benchmark. Our approach transforms the SLM into a minimalist agent utilizing two specialized tool calls: \texttt{extract\_facts} for symbolic triplet extraction and \texttt{get\_hint} for expert reasoning via a Relational Graph Convolutional Network (RGCN). We evaluate these models across two configurations, both in an \textit{Oracle} scenario with ground-truth triplets and a \textit{Realistic} scenario relying on self-extracted knowledge. Our results reveal that while \textbf{RGCN-derived hints provide a 1.5 - 2x performance gain} over story-only baselines, the system is constrained by the extraction bottleneck and sequential deductive fragility, where early extraction errors compound over multi-hop chains. Furthermore, we identify a ``distraction effect" in specific architectures where noisy, self-generated facts degrade performance despite the presence of expert hints. This work characterizes the challenges of symbolic grounding in low-resource agentic systems and provides a roadmap for iterative verification in neuro-symbolic agentic pipelines.
\end{abstract}


\maketitle




\section{Introduction}
Large Language Models (LLMs) have emerged as a standard for zero-shot reasoning across diverse domains \cite{llms1,llm22,llms3}. However, the deployment of frontier models (i.e., models with hundreds of billions of parameters) remains cost-prohibitive and has negative environmental impact \cite{environment,environment2}. Additionally, such models often require massive computational clusters that are inaccessible to the average user. Consequently, there is a growing shift toward Small Language Models (SLMs), which offer a more sustainable and budget-friendly alternative for low-resource environments \cite{slms1,slms2,slms3,slms4}.

\noindent Despite their efficiency, SLMs often struggle with complex, multi-hop relational reasoning where logical consistency is very important \cite{slms_scale, llm_abilities, llm_abilities1}. A notable example is the CLUTRR benchmark~\cite{sinha2019clutrr}, which requires models to induce kinship relations from narrative text through varying lengths of reasoning chains. In such scenarios, the limited parameter count of SLMs often leads to ``hallucinated" relations or a failure to maintain the symbolic state of the story \cite{slm_hallucinate, slm_hallucinate1}. The vulnerability of SLMs in these tasks stems from sequential deductive fragility: as the reasoning depth increases, minor errors in early fact extraction act as false premises, compounding through the chain and leading to catastrophic logical failures \cite{noise_compound, llm_abilities1}.

\noindent To address these limitations, we propose a neuro-symbolic framework that transforms the SLM from a standalone reasoner into a minimalist agent. Our approach augments the SLM with two specialized tool calls: (i) a fact extractor that generates relational triplets from the narrative (i.e., story), and (ii) an expert Relational Graph Convolutional Network (RGCN) \cite{rgcn} that reasons over the resulting knowledge graph to provide a ``hint" (i.e., a predicted relation and a confidence score) given a pair of query entities. By decoupling the text understanding from the relational logic, we provide a mechanism for causal grounding and structured explanation.

\noindent Our main contributions are as follows:
\begin{itemize}
\setlength\itemsep{.3em}
    \item We introduce a framework for enhancing SLM performance in zero-shot relational learning. In realistic settings where the SLM extracts its own facts, our method yields a 1.5 - 2x performance gain over story-only baselines.
    \item We provide a comparative study across different open-source model architectures (Gemma 3 1B/4B \cite{gemma}, Llama 3.2 3B \cite{llama}), analyzing how model size influences the quality of symbolic extraction.
    \item We conduct an extensive practical analysis, comparing our realistic pipeline against an Oracle configuration to isolate the failure modes inherent in noisy, self-generated knowledge graphs.
\end{itemize}

\section{Related Work}

\subsection{Relational Reasoning and CLUTRR}
Relational reasoning remains a cornerstone of artificial intelligence, requiring models to induce logical rules from discrete facts. The CLUTRR benchmark~\cite{sinha2019clutrr} was specifically designed to evaluate the systematic generalization of Natural Language Understanding (NLU) systems by requiring multi-hop kinship inference from story text. While graph-based models \cite{rgcn, gcn} operating on symbolic inputs often achieve high accuracy, standard language models exhibit a sharp performance decline as reasoning depth (i.e., hop count betweeen the query entities) increases, often due to an inability to maintain consistent symbolic state.

\subsection{Reasoning in Small Language Models (SLMs)}
Recent advancements have focused on narrowing the ``intelligence gap" between frontier models and SLMs through specialized training and high-quality synthetic data. Works such as \cite{slms_good, slms_good1} demonstrate that models under 4 billion parameters can perform competent single-hop tasks but struggle with the ``lost-in-the-middle" phenomenon~\cite{Liu2023} and logical chaining in zero-shot settings. Efforts to enhance SLMs often involve parameter-efficient fine-tuning (LoRA) or knowledge distillation from larger ``teacher" models to improve their logical grounding.\\
Recent research has identified a ``quantization trap" in small models, where the accumulation of noise across logical chains violates standard linear scaling laws, making multi-hop tasks uniquely difficult for compressed architectures \cite{slms_good2, slms_good1}. While multi-agent architectures have shown promise in maintaining stability across 4+ hops \cite{Sadeddine2025}, these solutions often re-introduce the computational costs we aim to avoid. Our work specifically targets the extraction phase, investigating whether specialized graph-based models can mitigate this ``sequential fragility" in a single-agent, tool-augmented setup.

\subsection{Neuro-Symbolic and Agentic Integration}
Neuro-symbolic AI seeks to combine the robust pattern matching of neural networks with the precision of symbolic logic. Recent ``agentic" frameworks have moved toward tool-augmented reasoning, where LLMs assign complex tasks, such as mathematical solving or structured planning, to specialized external modules \cite{nesy, nesy1}. Our work aligns with recent pipelines that utilize LLMs for triplet extraction to build dynamic knowledge graphs \cite{nlp_triplets}, though we specifically focus on the performance degradation introduced when SLMs are used as the primary extractors in noisy, low-resource scenarios.
\section{Methodology}\label{sec:method}

\begin{figure}[h]
    \centering
    \includegraphics[width=\textwidth]{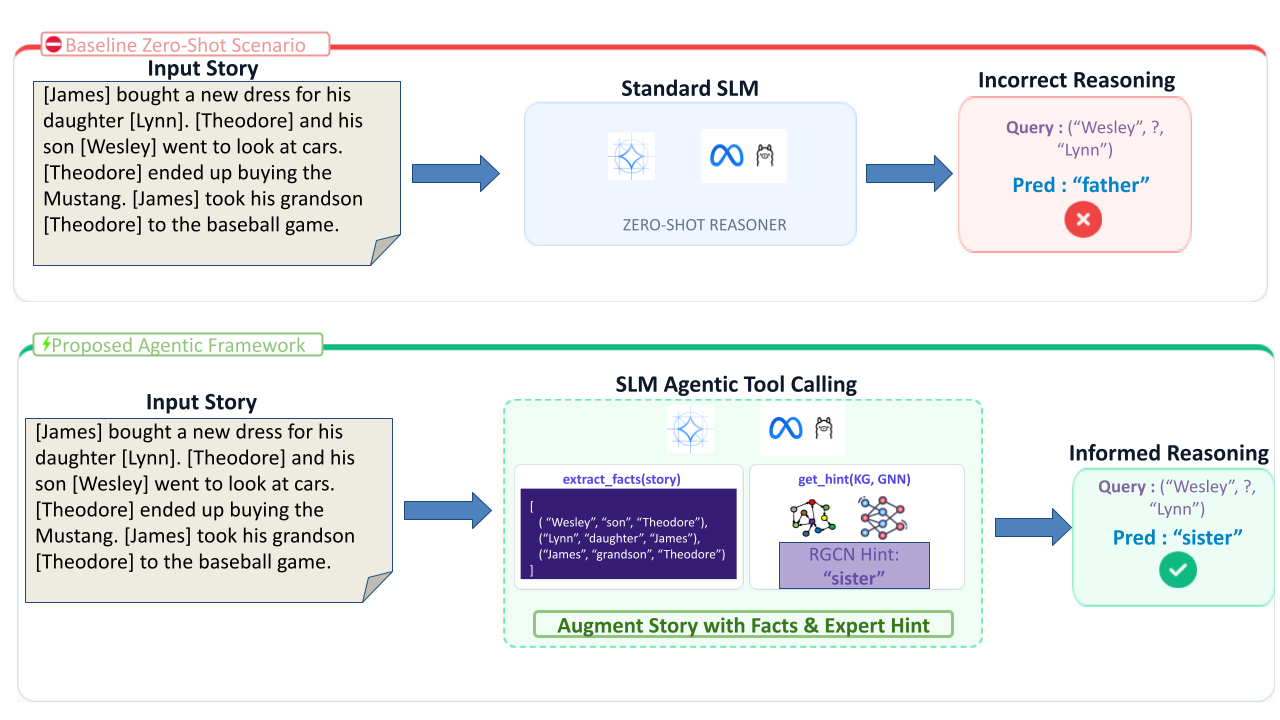}
    \caption{The proposed agentic loop. The SLM invokes \texttt{extract\_facts} to populate a per-story Knowledge Graph, which is then processed by the \texttt{get\_hint} tool to provide assistance for the final inference.}
    \label{fig:agentic_loop}
\end{figure}

Our framework (shown in Figure \ref{fig:agentic_loop}) adopts an agentic neuro-symbolic architecture where an SLM serves as the reasoning controller, assigning specialized tasks to symbolic and neural tools. This decoupling allows the model to ground its story understanding in explicit logical structures.

\subsection{Agentic Tool Calling Mechanism}
We define two primary tools that the SLM can invoke to enhance its zero-shot reasoning:
\begin{itemize}
    \item \textbf{\texttt{extract\_facts(story)}:} The SLM parses the story to extract kinship triplets $(u, rel, v)$. In our implementation, we follow a reversed edge direction where $(object, rel, subject)$, following the dataset’s logical schema.
    \item \textbf{\texttt{get\_hint(K\_graph, target\_pair)}:} This tool queries a pre-trained Relational Graph Convolutional Network (RGCN), with the query-based Knowledge Graph $K\_graph$ (generated by the
    extracted facts) as input.  The expert (i.e., the RGCN) returns the most probable kinship relation and a confidence score $\sigma$, which is then injected back into the SLM's context.
\end{itemize}
The \texttt{extract\_facts} tool utilizes a structured, zero-shot prompting strategy to distill story text into symbolic triplets. Given the constrained parameter count of our models, we enforce a strict output format to minimize parsing errors. The exact prompt template, including the kinship ontology and formatting constraints used to ensure consistent extraction across Gemma 3 and Llama 3.2, is provided in Appendix \ref{sec:prompts}.\\
Moreover, adhearing to the dataset's logical schema, we adopt a \textit{target-centric} representation where a triplet $(u, r, v)$ indicates that node $v$ is the $[r]$ of node $u$. For instance, in a father-daughter relationship, an edge from the father to the daughter is labeled as ``daughter", representing the relationship from the perspective of the target node. This reversed directionality is consistently maintained during the RGCN's training phase. By grounding the \texttt{get\_hint} tool in this specific directional logic, we ensure that the expert model's message-passing layers correctly aggregate the hierarchical context needed to resolve multi-hop queries (e.g., traversing multiple ``child" edges to infer a ``grandchild" relation). This precision prevents the ``logical inversion" errors common in standard LLM/SLM reasoning over complex family trees.

\subsection{Knowledge Graph Construction and Extraction Noise}
A core focus of this study is the realistic scenario, where triplets are extracted via zero-shot prompting rather than provided by an Oracle. This introduces \textit{extraction noise} in the form of hallucinated edges or missing relations. We characterize this as a challenge of sequential deductive fragility: in a multi-hop reasoning task, a single error in the early extraction of a kinship link (e.g., mistaking a ``sister" for a ``cousin") propagates through the graph, leading to a catastrophic failure in the RGCN's path-finding for high-hop queries. Additionally, missing a single kinship (i.e., a single relationship) might lead to disconnected nodes, hence making the subsequent task of kinship prediction by the RGCN more difficult (or in some cases infeasible).\\
To mitigate this phenomenon, we augment the extracted facts in a post-hoc manner by incorporating their corresponding inverse triplets. This approach ensures bidirectional information flow along the semantic chain, regardless of the original directionality of the facts as extracted by the SLM. Overall, the average recall of ground-truth facts per story was estimated to be approximately 71\%.

\subsection{RGCN Expert Training and Generalization}
The \texttt{get\_hint} tool is powered by an RGCN trained specifically on the training partition of the CLUTRR dataset. Crucially, the model is trained on stories with reasoning lengths of $k \in \{2, 3, 4\}$ hops. We then evaluate its performance on test sets containing up to $10$ hops. This setup tests the model's ability for systematic generalization, applying learned kinship rules to reasoning chains significantly longer (up to five times longer) than those encountered during training. By providing these GNN-derived hints to the SLM, we aim to compensate for the ``lost-in-the-middle" and context-window limitations typical of models in the 1B - 4B parameter range.
\section{Experiments and Discussion}\label{sec:experiments}

\subsection{Experimental Setup}
To evaluate the efficacy of our neuro-symbolic agentic framework, we standardize the training and inference parameters across all configurations. Our study utilizes three state-of-the-art open-weights models: Gemma 3 (1B, 4B) and Llama 3.2 (3B). The \texttt{extract\_facts} tool is implemented via a zero-shot prompting strategy, as detailed in Appendix \ref{sec:prompts}, while the expert reasoning is assigned to a Relational Graph Convolutional Network (RGCN).

\noindent The RGCN expert is configured with 4 layers to facilitate deep multi-hop message passing. We train the model for 20 epochs using the Adam optimizer with a learning rate of $10^{-3}$. Following the standard CLUTRR evaluation protocol, the RGCN is trained exclusively on the training partition containing reasoning chains of $k \in \{2, 3, 4\}$ hops. We then evaluate its zero-shot performance on test set, with
queries needing 2 to 10 hops to be answered, to measure the model's capacity for inductive logical generalization. All SLM inferences are conducted in a zero-shot setting to maintain the integrity of the agentic tool-calling paradigm without model-specific fine-tuning.

\subsection{Performance in the Oracle Scenario}
The quantitative results presented in Table \ref{tab:configuration_results} establish a strong contrast between theoretical reasoning capacity and practical zero-shot implementation. In the baseline  (i.e., zero-shot prediction using only the Story as input), SLMs struggle significantly; for instance, Gemma 1B fails to exceed a 7.54\% accuracy. This gap is dramatically bridged in the Oracle scenarios. When provided with a ground-truth \textit{Oracle Hint} (i.e., the actual relation to be predicted), Llama 3.2 3B experiences a nearly 4x performance increase, reaching 62.79\%. This increase suggests that the primary obstacle for SLMs is not a lack of linguistic comprehension, but rather an inability to maintain and query a consistent symbolic state over long, distracting contexts. Note that the last two rows in Table \ref{tab:configuration_results} correspond to the test accuracy of the RGCN, using Knowledge Graphs constructed by Oracle and SLM-extracted Facts respectively.

\begin{table}[htbp]
\centering
\caption{Model performance (measured in Accuracy \% ) across different prompt configurations.}
\label{tab:configuration_results}
\begin{tabular}{lccc}
\toprule
\textbf{Configuration} & \textbf{Gemma 1B} & \textbf{Llama 3B} & \textbf{Gemma 4B} \\
\midrule
Story & 7.54 & 16.13 & 13.45 \\
Oracle Facts & 7.73 & 19.08 & 15.84 \\
Story + Oracle Facts & 8.21 & 20.80 & 15.55 \\
Story + Oracle Hint & \textbf{34.16} & 62.79 & 59.16 \\
Oracle Facts + Oracle Hint & 28.24 & 61.83 & \textbf{74.43} \\
Story + Oracle Facts + Oracle Hint & 34.06 & \textbf{62.98} & 54.10 \\

\midrule
Story + SLM Facts & 6.01 & 16.61 & 1.64 \\
Story + GNN Hint (from SLM Facts) & \textbf{12.60} & 20.32 & \textbf{19.75} \\
Story + SLM Facts + GNN Hint (from SLM Facts) & 12.40 & \textbf{20.52} & 14.98 \\
\midrule
 & \multicolumn{3}{c}{\textbf{Expert Performance}} \\
RGCN on Oracle Facts & \multicolumn{3}{c}{60.69} \\
RGCN on SLM Facts & \multicolumn{3}{c}{24.88} \\
\bottomrule
\end{tabular}
\end{table}

\begin{figure}[htbp]
    \centering
    \includegraphics[width=\textwidth]{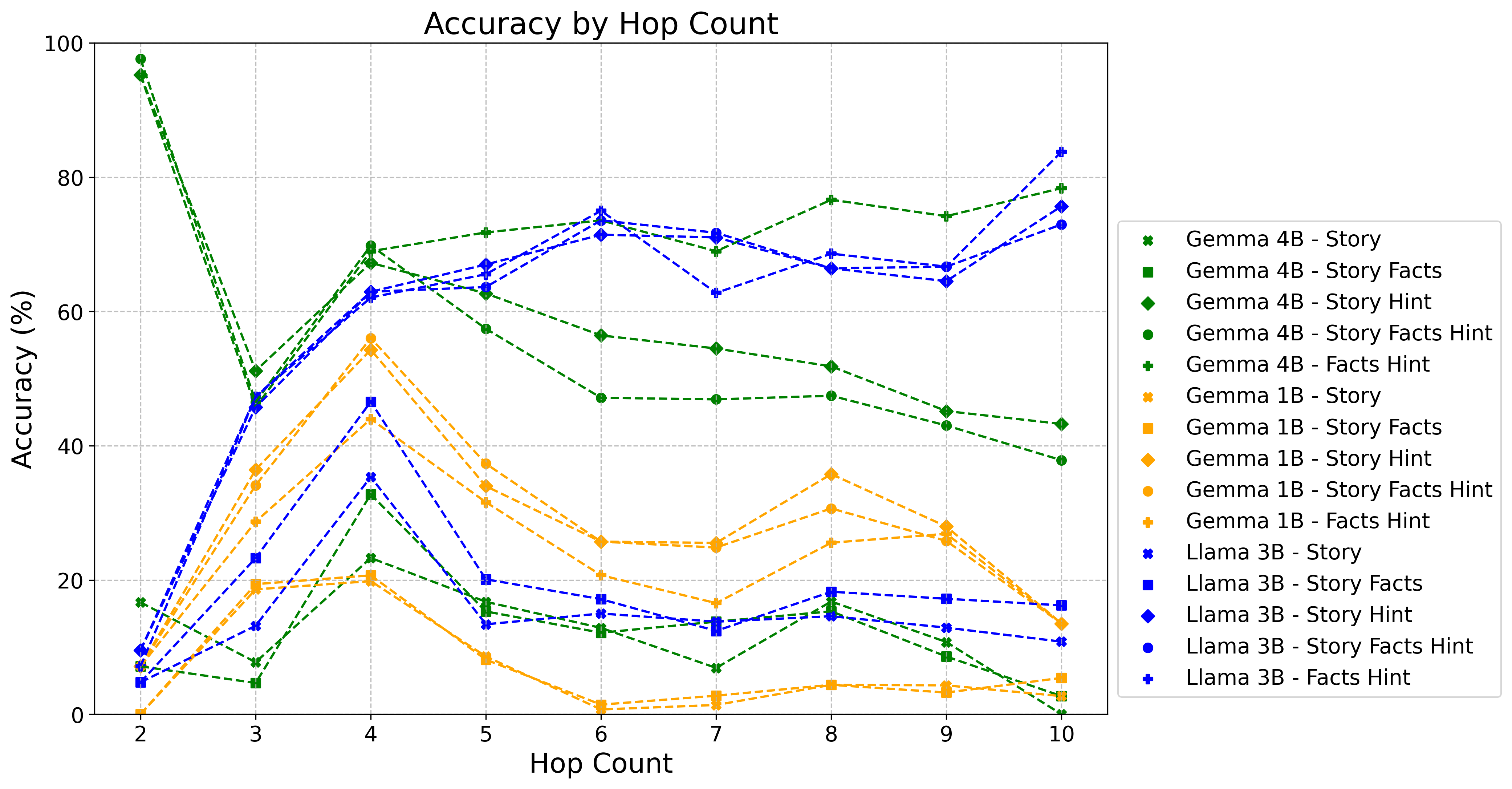}
    \caption{Model accuracy across queries with different hop-counts using Oracle-provided facts/hint. Performance tends to peak at hop counts up to 4, except mainly for the Llama-based models.}
    \label{fig:hop_oracle}
\end{figure}

\noindent In Figure \ref{fig:hop_oracle}, we illustrate the performance of each model and configuration, at each hop-level. Interestingly, the performance at the 2-hop baseline is not uniform across all architectures. While configurations utilizing the \textit{Oracle Hint} generally achieve high accuracy, only specific variants, most notably within the higher-parameter Gemma models (i.e., Gemma 4B), approach near-perfect scores at this stage. As the reasoning depth increases, a clear divergence in architectural resilience emerges. The Llama-based models exhibit better robustness; despite the training distribution for the expert reasoning being capped at 4 hops, these models maintain or even improve their accuracy as the chain extends toward 10 hops, particularly in configurations utilizing \textit{Oracle Hints}. This indicates both the capabilities of the RGCN expert in extracting relationships using multiple hops, even though it was trained in only up to 4 hops, and the efficiency of the SML in combining the expert hint with the \textit{Story}.

\noindent In contrast, the Gemma 3 variants, show a more noticeable and immediate performance decay as reasoning depth increases beyond the 4-hop threshold. This disparity suggests that the Llama architecture may possess a superior capacity for structured logical injection and long-range symbolic integration, whereas the Gemma models, particularly at the 1B scale, appear more susceptible to the ``Lost-in-the-Middle" phenomenon and contextual distraction in high-hop kinship queries, even when provided with high-quality guidance.

\subsection{Performance in the Realistic Scenario}
The transition to the Realistic scenario exposes the fragile nature of the neuro-symbolic loop. The RGCN expert, which achieves 60.69\% accuracy on ground-truth facts, sees its performance collapse to 24.88\% when operating on facts extracted by the SLMs (i.e., Gemma 4B). This drop confirms our hypothesis regarding sequential deductive fragility: because a multi-hop reasoning chain is only as strong as its weakest link, a single missed or hallucinated relation during the initial \texttt{extract\_facts} phase renders the resulting knowledge graph logically unrecoverable for the GNN solver.

\noindent An interesting finding in our study is that providing both facts and hints to the model, is not always helpful. As seen in Table \ref{tab:configuration_results}, for Gemma 4B, the configuration \textit{Story + GNN Hint} (19.75\%) notably outperforms \textit{Story + SLM Facts + GNN Hint} (14.98\%). This implies that the presence of noisy triplets acts as a ``logical anchor" that misleads the SLM, even when a correct expert hint is available in the prompt. Interestingly, Llama 3.2 3B exhibits greater resilience to this effect, maintaining nearly identical performance (20.32\% vs 20.52\%) regardless of the inclusion of noisy facts. This suggests that Llama's attention mechanism may be more efficient at filtering out symbolic noise in favor of high-confidence expert signals.

\begin{figure}[htbp]
    \centering
    \includegraphics[width=\textwidth]{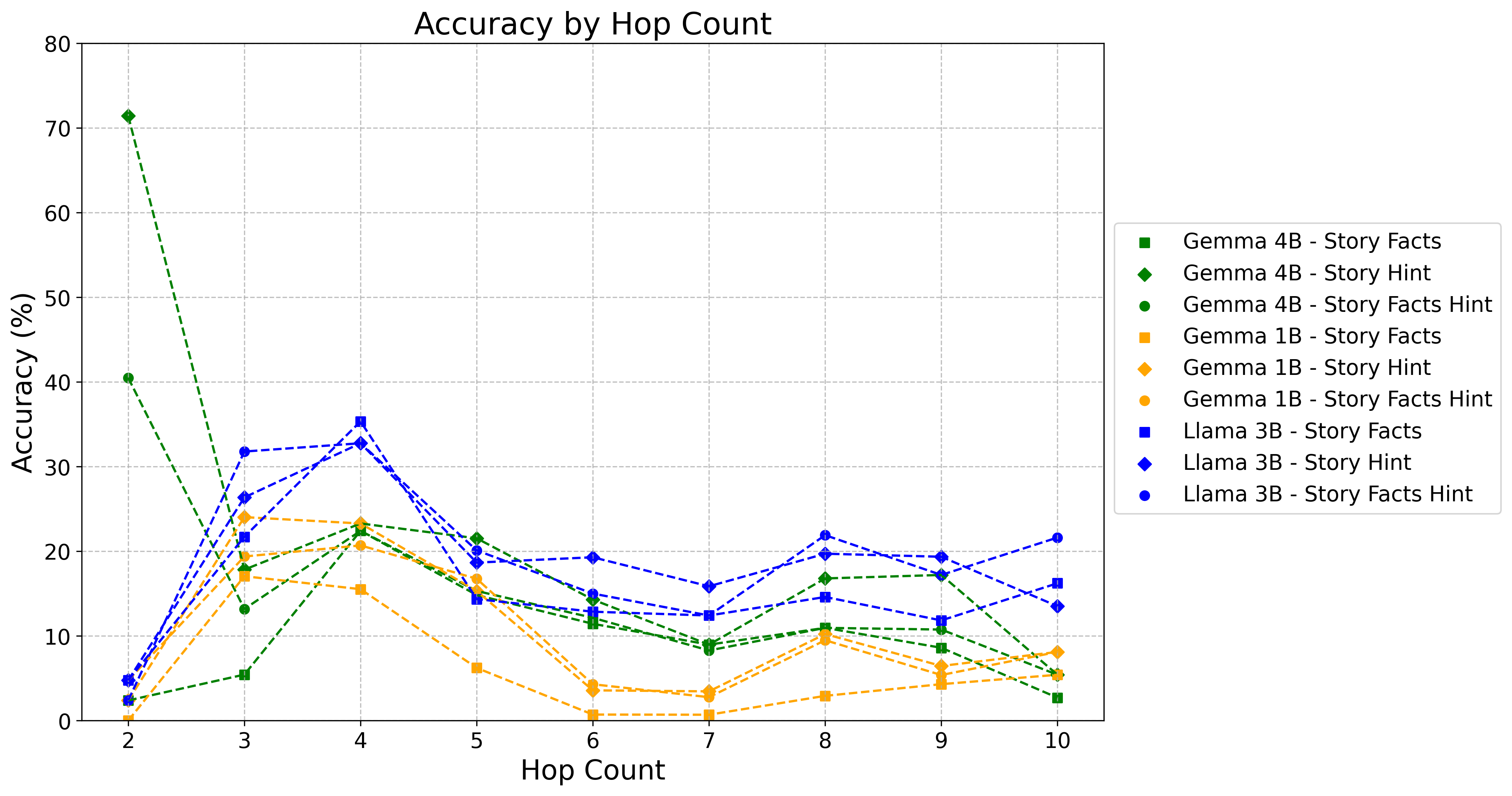}
    \caption{Model accuracy across queries with different hop-counts using SLM-generated facts/hint. Note the overall reduction in accuracy compared to Oracle-based configurations.}
    \label{fig:hop_llm}
\end{figure}

\noindent In Figure \ref{fig:hop_llm}, we illustrate the performance of the different configurations across varying hop levels, now using SLM-based facts/hint. A clear ``Reasoning Cliff" appears in this realistic scenario. While some oracle configurations maintain high accuracy after 4 hops, the realistic counterparts exhibit a catastrophic failure, performing with less than 20\% accuracy. Interestingly, most models struggle to exceed 5\% accuracy even for 2-hop chains. This indicates that the extraction process is most vulnerable at the point of initial fact grounding; if one of the few (two in this case) actual relations are not captured accurately, the GNN expert cannot provide a valid inference.

\noindent Performance generally peaks between 3 and 4 hops, where the relative density of the extracted graph may provide enough context for the GNN, before declining sharply. While the RGCN-derived hints provide a consistent 1.5 - 2x improvement over the story-only baseline for smaller models like Gemma 1B (moving from $\sim$7.5\% to 12.6\%), the gain remains insufficient to overcome the compounding noise of high-hop extractions. This suggests that the bottleneck is not the reasoning capacity of the GNN, but the quality of the symbolic grounding. 

\section{Conclusion}
In this work,  we investigated the performance of neuro-symbolic agentic systems on zero-shot relational reasoning, by evaluating the integration of Small Language Models (SLMs) with Relational Graph Convolutional Networks (RGCNs). Our investigation using the CLUTRR benchmark demonstrates that while SLMs like Llama 3.2 and Gemma 3 possess the latent capacity for complex relational reasoning when provided with high-quality symbolic guidance, they are significantly constrained by an extraction bottleneck. We observed that early errors in the information extraction phase compound over multi-hop chains, leading to a collapse in accuracy, when queries need more than 4 hops to be answered. Furthermore, our results reveal a notable ``Distraction Effect", where the inclusion of noisy, SLM-extracted facts can degrade performance even when expert hints are present in the context.\\
Future work will focus on implementing iterative symbolic refinement cycles to prune extraction hallucinations and investigating architecture-specific optimizations to better leverage the structural resilience observed in Llama-based models. Bridging this extraction gap remains essential for developing reliable, autonomous systems capable of causal grounding in open-domain environments.

\begin{acknowledgments}
  GCP resources were provided by the National Infrastructures for Research and Technology GRNET and funded by the EU Recovery and Resiliency Facility.
\end{acknowledgments}

\section*{Declaration on Generative AI}
During the preparation of this work, the authors used GPT-5, Gemini 3, and Grammarly for the purposes of: Drafting content, Grammar and spelling check, Paraphrase and reword, Improve writing style.\\ 
After using these tools/services, the authors reviewed and edited the content as needed and take full responsibility for the publication’s content. 

\bibliography{biblio}

\clearpage
\appendix

\section{Prompts used in Language Models}\label{sec:prompts}

The following table contains the prompts used for the different scenarios as explained in the Methodology Section~\ref{sec:method}. The available kinship relations appearing within the dataset are the following: aunt, brother, daughter, daughter-in-law, father, father-in-law, wife, husband, sister-in-law, granddaughter, grandfather, grandmother, grandson, mother, mother-in-law, nephew, niece, sister, son, son-in-law, uncle.

\begin{table}[ht]
\begin{small}

\centering
\caption{Different prompts used with SLMs. Preamble, Constraints, and Instructions are used as parts of subsequent prompt formulations.}\label{tab:zero_shot_prompts}
\begin{tabularx}{\linewidth}{cX}
\hline
\textbf{Variant} & \textbf{Prompt}\\
\hline
Preamble & {\ttfamily You are a Formal Logic Compiler. You process kinship data as a directed graph. IMPORTANT: Treat all human names as abstract node identifiers. Disregard gender norms, social tropes, or cultural assumptions associated with these names. Output ONLY valid JSON. 

Task: Determine the relationship between A and B.} \\
\hline
Constraints & {\ttfamily Constraints:

- Path tracing direction: Identify how A (Subject) is related to B (Object).

- The answer must be a single word from this list: <all available kinships>.}\\
\hline
Instructions & {\ttfamily Instructions:

1. Identify the path: B -> ... -> A.

2. Calculate the composition of relationships along that path.

3. Output the result as: {"answer": "term"}

4. Logical facts may contain false information.}\\
\hline
Story & Preamble + \texttt{Story: <story\_text>} + Constraints + Instructions\\
\hline
Story + Facts & Preamble + \texttt{Story: <story\_text>} + \texttt{Logical Facts: <logical\_facts>} + Constraints + Instructions \\
\hline
Story + Hint & Preamble + \texttt{Story: <story\_text>} + Constraints + \texttt{Hint: <expert\_hint>} + Instructions \\
\hline
Facts + Hint & Preamble + \texttt{Logical Facts: <logical\_facts>} + Constraints + \texttt{Hint: <expert\_hint>} + Instructions \\
\hline
Story + Facts + Hint & Preamble + \texttt{Story: <story\_text>} + \texttt{Logical Facts: <logical\_facts>} + Constraints + \texttt{Hint: <expert\_hint>} + Instructions \\
\hline

\hline
\end{tabularx}
\end{small}
\end{table}

\noindent Regarding the \texttt{extract\_facts} procedure for Gemma 4B we use the following prompt:

\begin{lstlisting}[style=mystyle]
"""    
You are an expert Information Extraction system.

Task: Extract all explicit kinship relationships from the short story below.
    
Story: "{story}"
    
    Instructions:
    1. Identify people names enclosed in brackets (e.g., [Alice]). Extract them WITHOUT brackets.
    2. Identify the relationship between them.
    3. IMPORTANT: Extract relations in an ASCENDANT fashion (Child -> Parent). E.g.: 
       - Use grandson/granddaughter instead of grandfather/grandmother.
       - Use son/daughter instead of father/mother.
     etc.
"""
\end{lstlisting}

\end{document}